\documentclass[runningheads]{llncs}
\usepackage{graphicx}
\usepackage{amsmath,amssymb} 
\usepackage{color}

\usepackage{graphicx}
\usepackage{amssymb}
\usepackage{times}
\usepackage{amsmath}

\usepackage{epsfig}
\usepackage{multirow}
\usepackage{tabularx}
\usepackage{cancel}
\usepackage{epstopdf}
\usepackage{arydshln}
\usepackage{color,ulem}
\usepackage{url}
\usepackage{subfig}
\usepackage{pbox}

\begin{document}

\newcommand{\point}{
    \raise0.7ex\hbox{.}
    }


\pagestyle{headings}

\mainmatter

\title{Calorie Counter: RGB-Depth Visual Estimation of Energy Expenditure at Home} 

\titlerunning{RGB-Depth Visual Estimation of Energy Expenditure} 

\authorrunning{Tao et.al.} 

\author{Lili Tao, Tilo Burghardt, Majid Mirmehdi, Dima Damen, Ashley Cooper, \\ Sion Hannuna, Massimo Camplani, Adeline Paiement, Ian Craddock} 
\institute{SPHERE, Faculty of Engineering, University of Bristol} 

\maketitle

\begin{abstract}
We present a new framework for vision-based estimation of calorific expenditure from RGB-D data - the first that is {validated} on physical gas exchange measurements and {applied} to daily living scenarios. Deriving a person's energy expenditure from sensors is an important tool in tracking physical activity levels for health and lifestyle monitoring. Most existing methods use metabolic lookup tables~(METs) for a manual estimate or systems with inertial sensors which ultimately require users to wear devices.  In contrast, the proposed pose-invariant and individual-independent vision framework allows for a remote estimation of calorific expenditure. {We introduce, and evaluate our approach on, a new dataset called \textit{{SPHERE-calorie}}, for which visual estimates can be compared against simultaneously obtained,} indirect calorimetry measures based on gas exchange.
We conclude from our experiments that the proposed vision pipeline is suitable for home monitoring in a controlled environment, with calorific expenditure estimates above accuracy levels of commonly used manual estimations via METs.
With the dataset released, our work establishes a baseline for future research for this {little-explored area} of computer vision. 
\end{abstract}


\section{Introduction}
\label{sec:intro}

The large majority of research into the physical activity levels of people with, or at risk of, chronic disease have measured either total physical activity or physical activity acquired in specific activities, such as walking, that generally occur outside the home. Very little is known about how activities of normal daily living in the home environment may contribute to prevention of or recovery from/management of chronic disease, such as obesity and diabetes. An accurate assessment of physical activity within the home is thus important to understand recovery progress and long term health monitoring \cite{samitz2011}.

Energy expenditure, also referred to as `calorific expenditure', is one commonly used single metric to quantify physical activity levels over time. It provides  a key tool for the assessment of physical activity; be that for the long term monitoring  of  health and lifestyle aspects associated to chronic conditions or for recovery medicine. Calorific expenditure is traditionally  measured either using direct methods, such as a sealed respiratory chamber \cite{ravussin1986determinants}, or indirect calorimetry, which requires carrying gas sensors and wearing a breathing mask \cite{K4b2}. The latter is often based on the respiratory differences of oxygen and carbon dioxide in the inhaled and exhaled air. It forms the measurement standard for non-stationary scenarios where the person can move freely. Recently, the use of wearable devices -- with a focus on coarse categorisations of activity levels by wrist-worn inertial sensors \cite{altini2015estimating} -- has become a popular monitoring choice due to its low cost, low energy consumption, and data simplicity. Among these, tri-axial accelerometers are the most broadly used inertial sensors \cite{chen2015improving}. 

Visual systems, on the other hand, are already a key part of home entertainment systems today, and their RGB-D sensors \cite{aggarwal2014human} allow for a rich and fine-grained analysis of human activity within the field of view. Recent advances in computer vision have now opened up the possibility of integrating these devices seamlessly into home monitoring systems~\cite{Zhu_etal_2015,Woznowski_etal_2015}. 

\begin{figure*} [t]
\centering
\includegraphics[width=0.95\textwidth]{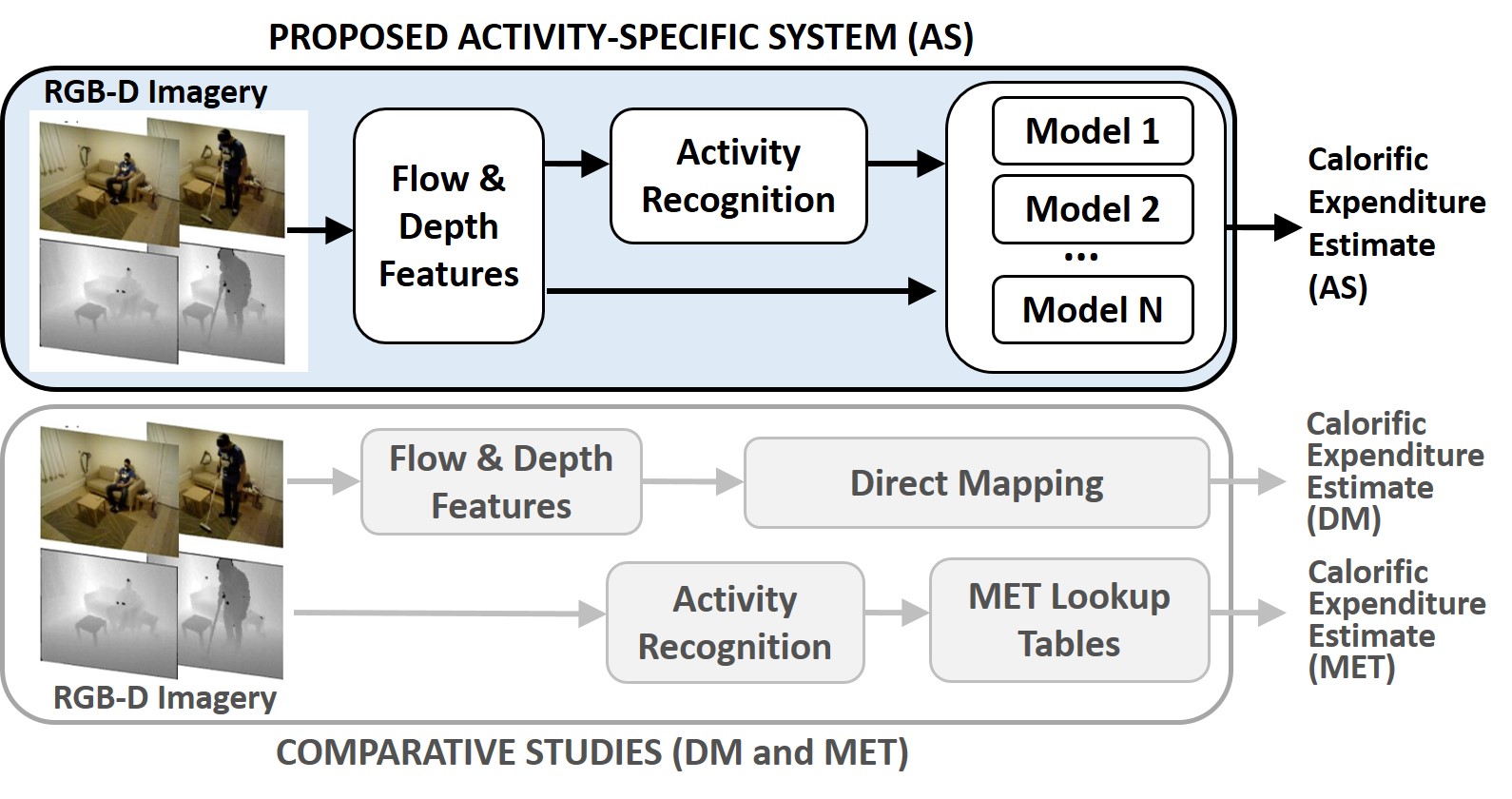}\vspace{-10pt}
\caption{Framework Overview. RGB-D videos are represented by a combination of flow and depth features. The proposed recurrent method AS (top) then selects activity-specific models which map to energy expenditure estimates. We compare this method to a direct mapping method DM and a manual estimate via lookup tables MET (bottom).\vspace{-10pt}
}
\label{Fig:models}
\end{figure*}

With this in mind, we propose a framework for estimating energy expenditure from RGB-D data in a living room environment. Figure \ref{Fig:models} shows in bold a flowchart of the proposed method -- mapping visual flow and depth features to calorie estimates using activity-specific models (AS in short). The method implements a cascaded and recurrent approach, which explicitly detects activities as an intermediate to select type specific mapping functions for final calorific estimation. We compare this proposed method against a ground truth of gas-exchange measurements (GT in short) and two off-the-shelf alternatives: (1)~mapping features directly to calorie estimates via a monolithic classifier (DM in short), and (2) manual mapping from activity classes to calorie estimates via metabolic equivalent task lookup tables~\cite{ainsworth2000compendium}~(MET in short) as most often applied in practice today. 

This is a new application in computer vision where no existing datasets are available. In order to quantify the performance, we introduce {a new dataset, \textit{SPHERE-calorie},} for calorific expenditure estimation collected within a home environment
 The dataset contains 11 common household activities performed over up to 20 sessions, lasting up to 30 minutes for each session, in each of which the activities are performed continuously. The setup consists of an {RGB-D} Asus {Xtion} camera mounted at the corner of a living room and a COSMED K4b2 \cite{K4b2} indirect calorimeter for ground truth measurement. The \textit{SPHERE-calorie} dataset will be publicly released.
In summary, the major contributions of this paper {are, (a) a first-ever} framework for a vision-based estimation of calorific expenditure from RGB-D data only, applicable to daily living scenarios, {and (b)} a novel dataset linking more than 10 hours of RGB-D video data  to  ground truth calorie readings from indirect calorimetry based on gas exchange.

\section{Related Work}
Applying computer vision techniques to help with the diagnosis and management of health and wellbeing conditions has gained significant momentum over the last years. However, studies on energy expenditure using visual sensors have been relatively limited. Our work explores this field further and builds on several relevant subject areas in vision. 

\textbf{Feature Representation -}
The visual trace of human activity in video forms a spatio-temporal pattern. To extract relevant properties from this for the task at hand, one aims at  compactly capturing this pattern and highlighting important aspects related to the properties of interest. Assuming that both body configuration and body motion~\cite{guo2014survey} are relevant to infer calorific uptake, the pool of potential features is large - ranging from local interest point configurations \cite{laptev2005space}, over holistic approaches like histograms of oriented gradients and histograms of motion information \cite{Tao_etal_2015}, to convolutional neural network features \cite{jia2014caffe}. 

Motion information in the first place could also be recovered in various ways, e.g. from RGB data using optical flow \textit{or} from depth data using 4D surface normals \cite{oreifej2013hon4d}. Whilst a composition of these features via concatenation of per-frame descriptors is straight forward, this approach suffers from the curse of dimensionality and unaffordable computational cost. Sliding window methods~\cite{Tao_etal_2016}, on the other hand, can limit this by predicting current values only from nearby data within a temporal window.
Further compaction may be achieved by converting large feature arrays into a single, smaller vector with a more tractable dimension count via, for instance, bags of visual words~\cite{laptev2008learning}, Fisher vectors\cite{perronnin2010improving} or time series pooling~\cite{ryoo2015pooled}. {In summary, the challenge of feature representation will require capturing visual aspects relevant to calorific expenditure, whilst limiting the dimensionality of the descriptor.}

\begin{figure*} 
\centering
\includegraphics[width=1\textwidth]{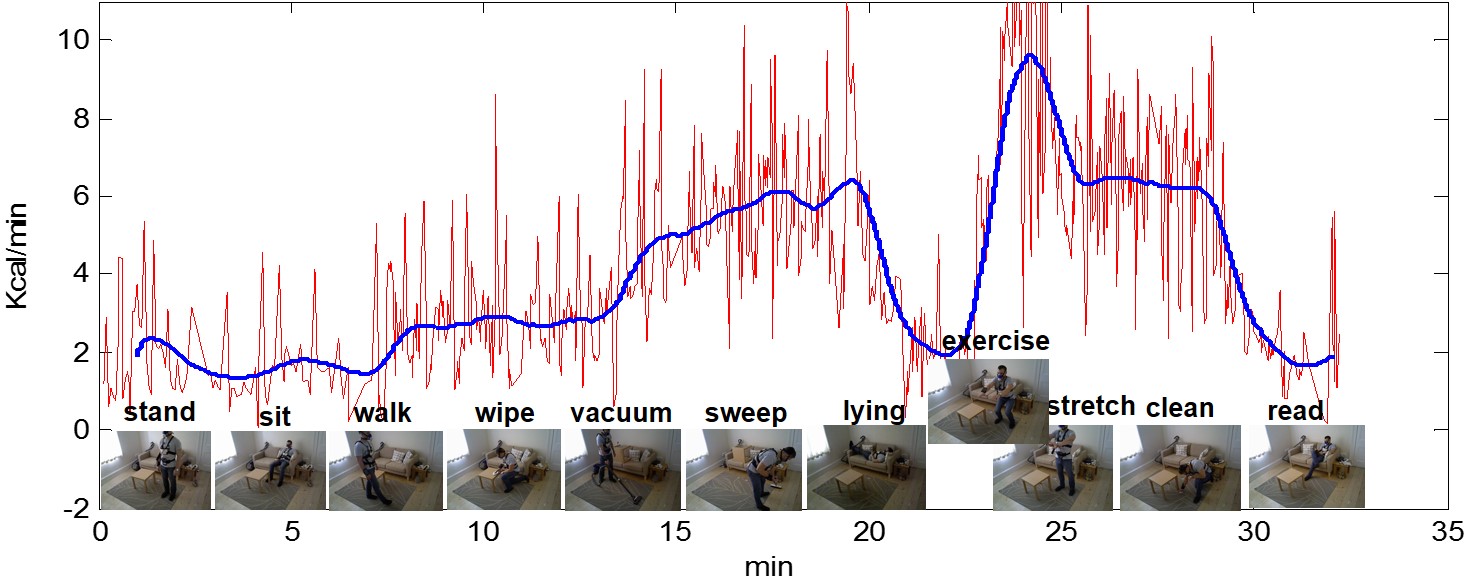}\vspace{-8pt}
\caption{\textbf{Ground truth example sequence.} Raw per breath data (red) and smoothed  COSMED-K4b2 calorimeter readings (blue) and sample colour images corresponding to the activities performed by the subject.\vspace{-10pt}
}
\label{Fig:Line}
\end{figure*}

\smallbreak
\textbf{Activity Recognition - }
There exists a significant body of literature describing the inference of activities from 2D colour intensity imagery~\cite{aggarwal2011human}, RGB-D data~\cite{aggarwal2014human}, and skeleton-based data \cite{presti20163d}. {Knowledge about the type of activity undertaken has been shown to correlate with the calorific expenditure incurred \cite{ainsworth2000compendium}. In alignment with Figure~\ref{Fig:models}, we will argue in this work that an explicit activity recognition step in the vision pipeline can, as an intermediate component, aid the visual estimate of energy uptake.}   

\smallbreak
\textbf{Visual Energy Expenditure Estimation -}
2D video has recently been used by Edgcomb and Vahid~\cite{edgcomb2013estimating} coarsely to estimate daily energy expenditure. In their work a subject is segmented from the scene background.  {Changes in height and width of the subject's motion bounding box, together with vertical and horizontal velocities and accelerations,} are then used to estimate calorific uptake. 
{Tsou and Wu \cite{tsou2015estimation} take this idea further and estimate  calorie consumption using full 3D joint movements tracked as skeleton models by a Microsoft Kinect.}
Both of the above methods use wearable accelerometry as the target ground truth, which in fact does not provide an accurate benchmark; and skeleton data is commonly noisy and currently only operates reliably when the subject is facing the camera. This limits applicability in more complex visual settings as contained in the \textit{SPHERE-calorie} dataset.

As outlined in the following section, our work attempts to remedy these shortcomings by using skeleton-independent, RGB-D based vision to estimate calorific expenditure against a standardised   calorimetry sensor  COSMED-K4b2 based on gas exchange~(see~Figure~\ref{Fig:Line}).

\section{Proposed Method} \label{Sec:method}

We propose an activity-specific  pipeline to estimate energy expenditure  utilising both depth and motion features as input. Importantly, our setup as shown in Figure~\ref{Fig:models} is designed to reason about activities \textit{first}, before estimating calorie expenditure via a set of  models which are each separately trained for particular activities.\vspace{-10pt}
\subsection{Features}\vspace{-5pt}
We first simultaneously collect RGB and depth imagery using an Asus Xmotion. For each frame~$t$, appearance and motion features are extracted, with the latter being computed with respect to the previous frame (level 0). A set of temporal filters is then applied to form higher level motion features~(level~1).  
We extract {the features within the bounding box returned by the OpenNI SDK \cite{OpenNI2010} person detector and tracker}. To normalise the utilised image region due to varying heights of the subjects and their distance to the camera, the bounding box is scaled by fixing its longer side to $M=60$ {pixels}, {a size  recognised as optimal for human action recognition~\cite{tran2008human},} while maintaining aspect ratio. The scaled bounding box is then centred in a $M \times M$ square box and horizontally padded.

\begin{figure*}[t]
\centering
\includegraphics[width=1\textwidth]{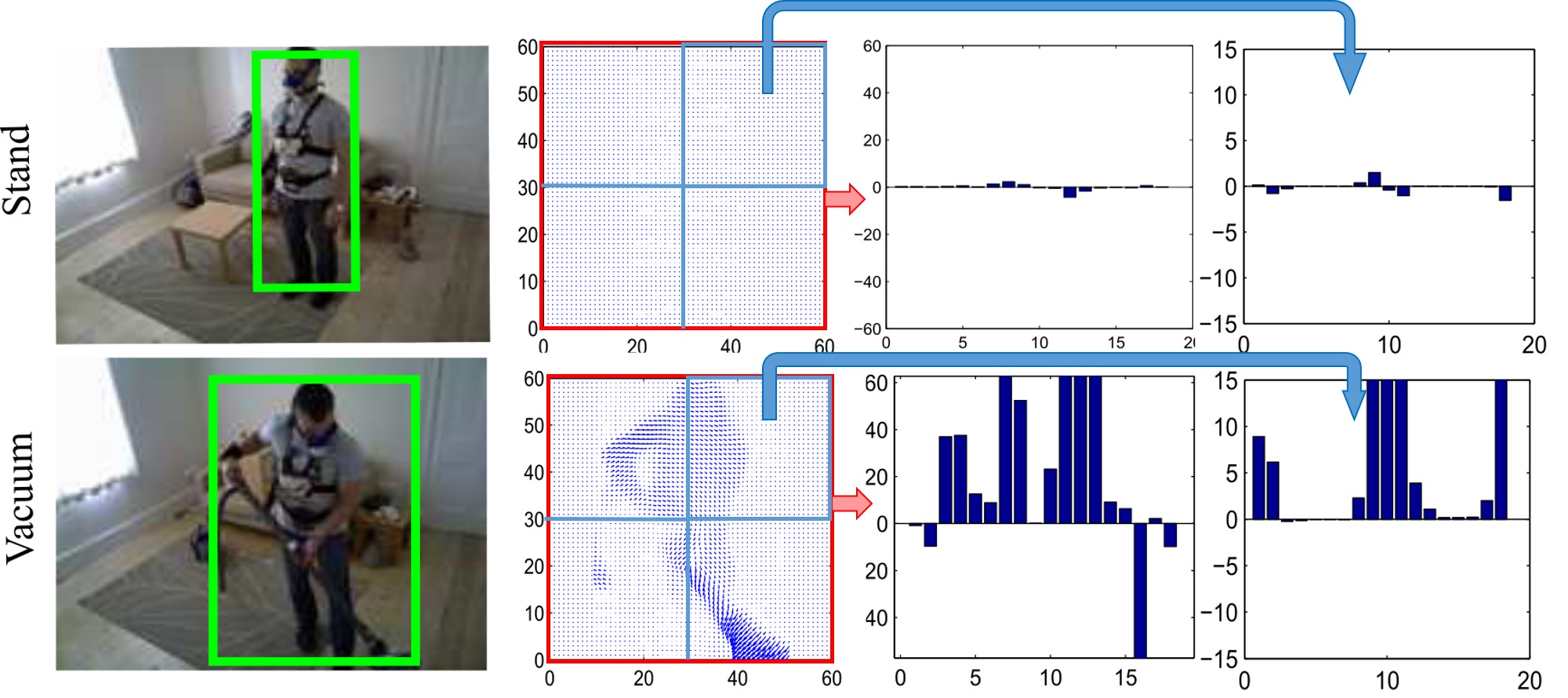}
\caption{{\textbf{Flow feature encoding via spatial pyramids.} \textit{Top row:} }limited motion while standing still. {\textit{Bottom row:}} significant motion features when moving during vacuuming. \textit{First column:} colour images with detected person. \textit{Second column:} optical flow patterns. \textit{Third column:} motion features at level 0. \textit{Last column:} motion features from the top-right quadrants of the image at level 1 (at which the image is subdivided into four quadrants).\vspace{-10pt}}
\label{Fig:color_flow_hist}
\end{figure*}

\noindent
\textbf{Motion feature encoding - } 
Inspired by \cite{tran2008human}, optical flow measurements are taken over the bounding box area and split into horizontal and vertical components. These are re-sampled to fit the normalised box and a median filter with kernel size $5 \times 5$ is  applied to smooth the data. A spatial pyramid structure is used to form hierarchical features from this. Such partitioning of the image  into an iteratively growing number of sub-regions increases discriminative power. The normalised bounding box is divided into a $n_g \times n_g$ non-overlapping grid, where~$n_g$ depends on {the} pyramid level, and the orientations of each grid cell are quantised into $n_b$ bins. The parameters for our experiments are empirically determined as $n_b = 9$ and $n_g=1$ and $2$ for levels 0 and 1 {respectively}. Figure \ref{Fig:color_flow_hist} exemplifies optical flow patterns and their encoding for different activities.   

\noindent
\textbf{Appearance feature encoding - }
{We extract depth features 
by applying the histogram of oriented gradients (HOG) feature on raw depth images \cite{dalal2005histograms} within the normalised bounding box. We then apply Principal Component Analysis (PCA) and keep the first 150 dimensions of this high-dimensional descriptor which retains $95\%$ of the total variance.} 

\noindent
\textbf{Pyramidal temporal pooling -} 
Given {the motion and appearance features extracted} from each frame in a sequence of images, it is important to capture both short and long term temporal changes and summarise them to represent the motion in the video. Pooled motion features were first presented in \cite{ryoo2015pooled}, designed for egocentric video analysis. We modify the pooling operator to make it more suitable for our data as follows.

\begin{figure*} [t]
\centering
\includegraphics[width=0.7\textwidth]{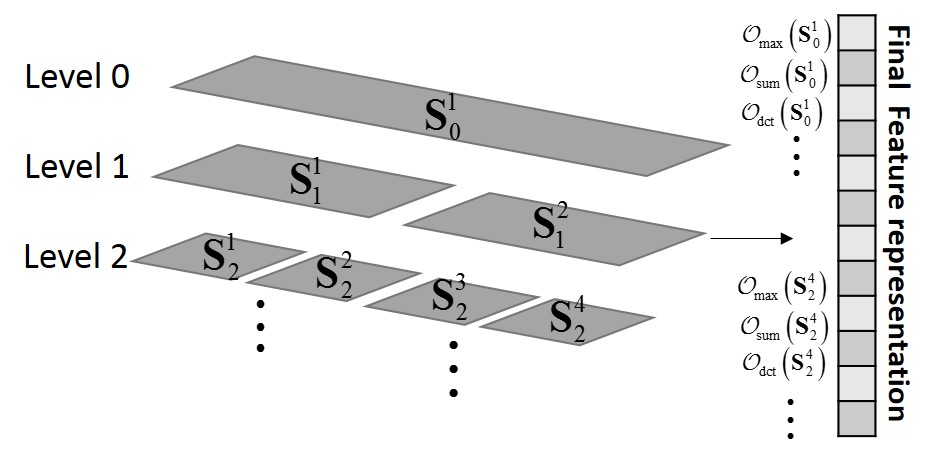}
\caption{\textbf{Temporal pyramid pooling and its feature representation.} This schematic shows the temporal subdivision of  data into various pyramidal levels (left) and the  concatenation of resulting feature (e.g. max, sum and dict) into a descriptor vector (right).\vspace{-10pt}}
\label{Fig:features}
\end{figure*}

An illustration of the temporal pyramid structure and the process for pooling operations are shown in Figure \ref{Fig:features}. The time series data $\mathbf{S}$ can be represented as a set of time segments at level $i$ as $\mathbf{S} = [\mathbf{S}_{i}^{1},\ldots,\mathbf{S}_{i}^{2^i}]$. The final feature representation is a concatenation of multiple pooling operators applied to each time segments at each level. The time series data can also be explained as $T$ number of per-frame feature vector,
such that $\mathbf{S} = \left\{S_1,\ldots ,S_N \right\}$, $\mathbf{S}\in {{\mathbb{R}}^{N\times T}}$ for a video in matrix form, where $N$ is the length of the {per-frame} feature vector, and $T$ is the number of frames. A time series $S_n = [ s_n(1),\ldots ,s_n(T)]$ is the $n^{th}$ feature across $1,\ldots ,T$ frames, where $s_n(t)$ denotes $n^{th}$ feature at frame $t$. 
A set of temporal filters with multiple pooling operators is applied to each time segment $[t_{min}, t_{max}]$ and produces a single feature vector for each segment via concatenation. 
We use two conventional pooling operators, max pooling and sum pooling, {as well as} frequency domain pooling. They are defined {respectively as}:
\vspace{-10pt} 
\begin{equation}
{\mathcal{O}_{\max }}({{S}_{n}})=\underset{t={{t}_{\min }}\cdots {{t}_{\max }}}{\mathop{\max }}\,{{s}_{n}}(t)
\mbox{~~~~~~~ and ~~~~~~~}
{\mathcal{O}_{\text{sum} }}({{S}_{n}})=\sum\limits_{{t=t_{\min}}}^{{t_{max}}}{{{s}_{n}}(t)}
\end{equation} 
Frequency domain pooling is used to represent the time series $S_n$ in the frequency domain by the discrete cosine transform, where the pooling operator {takes the} absolute value of the $j$ lowest frequency components of the frequency coefficients $D$,\vspace{-3pt} 
\begin{equation}
{\mathcal{O}_{\text{dct} }}({{S}_{n}})= \left| {{M}_{1:j}S_n} \right|
\vspace{-2pt}
\end{equation}
where $M$ is the discrete cosine transformation matrix.

\subsection{Learning and Recurrency}
\label{Sec:learning}
Energy expenditure estimation can be formulated as a sequential and supervised regression problem. We train a support vector regressor to predict calorie values from given features over a training set. 
The sliding window method is used to map each input window of width~$w$ to an individual output value $y_t$. The window contains the current and the previous $w-1$ observations. The window feature is represented by temporal pooling from the time series $\mathbf{S} = \left\{S_{t-w+1},\ldots ,S_t \right\}$. 

{We note that energy values for a particular time are highly dependent on the energy expenditure history. 
In our system, these are most directly expressed by previous calorific predictions during operation. Thus, employing recurrent sliding windows offers an option to not only use the features within a window, but also take the most recent $d$ predictions $\left\{\hat{y}_{t-d},\ldots, \hat{y}_{t-1}\right\}$ into consideration to help predict $y_t$. }During learning, as suggested in \cite{dietterich2002machine}, the ground truth labels in the training set are used in place of recurrent values.

\section{Experimental Results}

We introduce the RGB-D \textit{{SPHERE-calorie} dataset} captured in a real living environment. 
The ground truth was captured by {the COSMED} K4b2 portable metabolic measurement system. The dataset was generated over 20 sessions by 10 subjects with varying anthropometric {measurements} containing up to 11 activity categories per session, {and totalling around 10~hours recording time. 
{The categories and their associated MET values (in brackets) are: light intensity activities = \textit{\{sit still (1.3), stand still (1.3), lying down (1.3), reading (1.5)\}}; light+ intensity activities =  \textit{\{walking (2.0), wiping table (2.3), cleaning floor stain (3.0)\}}; moderate vigorous intensity activities =  \textit{\{vacuuming (3.3), sweeping floor (3.3), squatting (5.0), upper body exercise (4.0)\}}}. 

Colour and depth images were acquired at a rate of 30Hz. The calorimeter gives readings per breath, which {occurs} approximately every 3 seconds. To better model transitions between activity levels, we consider 9 different combinations of {the above three} activity intensities in each session.
Figure \ref{Fig:Line} shows a detailed example of calorimeter readings and associated sample RGB images from the dataset. The raw breath data is noisy (in red), and so we apply an average filter with a span of approximately $20$~breaths~(in blue). The participants were asked to perform the activities based on their own living habits without any extra instructions.

{We compare the proposed method AS to the direct mapping method DM and the metabolic equivalent table method MET. } {DM is formalised as $Y_t = f(X_t)$, where $Y_t$ is the target calorie value regardless of activity at time $t$, and $X_t$ contains the associated feature vector over a window. The goal is to find a function $f(\centerdot)$ that best predicts $Y_t$ from training data $X_t$}. 
{MET, widely used by clinicians and physiotherapists, assumes $N$ clusters of activity $\mathbf{A}=\left\{ {{A}_{1}},{{A}_{2}},\ldots ,{{A}_{N}} \right\}$  are known. A MET value is assigned to each cluster, together with anthropometric characteristics of individuals. The amount of activity-specific energy expended can then be estimated as $energy = 0.0175 (kcal/kg/min) \times weight (kg) \times$ MET values \cite{ainsworth2000compendium}. }
\vspace{-8pt}
\subsection{Evaluation and Parameter Settings}
In our experiments, we use non-linear SVMs with Radial Basis Function (RBF) kernels for activity classification and a linear support vector regressor for energy expenditure prediction. The libsvm \cite{chang2011libsvm}  implementation was used in the experiments. We perform a grid search algorithm to estimate the hyper-parameters of the SVM. For testing, we implement leave-one-subject-out cross validation on the dataset. This process iterates through all subjects, and the average testing error and standard deviation of all iterations are reported. We use the root-mean-squared error (RMSE) as a standard evaluation metric for the deviation of estimated calorie values from the ground truth. 
\subsection{Quantitative Evaluation} \vspace{-5pt}
{\bf Temporal Window Size -} The accuracy of predicted calorie values is linked to the number of previous frames utilised for making the prediction. The test described in this section looks at the relation between window length on the one hand, and activity recognition and calorie prediction errors on the other. All the sequences are tested with various window {frame lengths $w = \{450,900,1800\}$}, corresponding to a 15, 30 and 60 seconds time slot. Table \ref{Tab:window} illustrates the activity recognition rates and the average RMSEs for calorie prediction of different window length $w$.

{In general, the best performance for recognising activities is achieved when a 15~seconds window is applied. This is particularly prominent for individually highly variable activity types. For example, the recognition rates for exercise and stretch are significantly lower when $w=1800$. In these cases, data are likely to be better explained at a relatively small temporal interval, for which local temporal information are more descriptive.} On the contrary, calorie values are better predicted using {larger} window sizes. Here, human body adaptation causes an exponential increase/decrease to a plateau in oxygen consumption until a steady state corresponding to the current activity is attained~\cite{mcardle1991exercise}.

So far, we have applied the same window length for the prediction of calories and the detection of activities. In order to test how the estimated calorie value is influenced by the performance of action recognition, the proposed method is also tested with fixed window length $w=1800$ for predicting calorific expenditure, whilst different window lengths are applied to achieve different activity recognition rates. We also compare this to an idealised case by assuming all the activities are correctly recognised. For better visualisation, the 11 actions are grouped into three clusters based on {their} intensity level {in Figure \ref{fig:window_and_models}(a) which summarises the calorie prediction error} for different intensities and action recognition rates. We use normalised RMSE to facilitate the comparison between data with different scales.\vspace{-10pt}

\begin{table*} 
\footnotesize
\begin{center}
\begin{tabular}{|c|c|c|c|c|c|c|c|c|c|c|c|c|c|}
\hline 
 & {\bf w} & \textit{stand} & \textit{sit} & \textit{walk} & \textit{wipe} & \textit{vacuum} & \textit{sweep} & \textit{lying} & \textit{exercise} & \textit{stretch} & \textit{clean} & \textit{read} & {\bf overall}\tabularnewline
\hline 
\hline 
\multirow{3}{*}{\bf activity} & 450 &  \textbf{86.5}    &77.6   &88.3   &69.4   &79.0   &76.5   &\textbf{62.3}  &39.2   &\textbf{61.1}  &\textbf{91.4}  &\textbf{38.9}  &\textbf{73.7} \tabularnewline
\cline{2-14} 
 & 900 & 85.0   &79.1   &\textbf{89.4}  &\textbf{71.9}  &\textbf{81.1}  &75.2   &54.3   &\textbf{40.3}  &57.8   &90.4   &36.8   &{71.1} \tabularnewline
\cline{2-14} 
 & 1800 & 81.1  &\textbf{79.7}  &85.1   &66.0   &77.2   &72.9   &33.0   &29.3   &52.7   &90.0   &35.9   &{68.2} \tabularnewline
\hline \hline
\multirow{3}{*}{\bf calorie} & 450 &  1.41      &1.12   &0.76   &1.23   &1.19   &1.63   &1.95   &3.37   &2.91   &1.57   &1.68   &{1.55} \tabularnewline
\cline{2-14} 
 & 900 &  1.25  &0.87   &\textbf{0.76}  &\textbf{1.09}  &1.26   &1.47   &1.75   &2.82   &2.91   &\textbf{1.46}  &1.42   &{1.41} \tabularnewline
\cline{2-14} 
 & 1800 & \textbf{0.92} &\textbf{0.76}  &0.82   &1.17   &\textbf{1.19}  &\textbf{1.40}  &\textbf{1.54}  &\textbf{2.16}  &\textbf{2.81}  &1.49   &\textbf{1.32}  &\textbf{1.31}\tabularnewline
\hline 
\end{tabular}
\caption{Activity recognition rate (\%) and calorific expenditure prediction error (RMSE) with different window length. The best results in each activity are in bold.\vspace{-30pt}}
\label{Tab:window}
\end{center}
\end{table*}

\begin{figure} [ht] 
\centering
\subfloat[]{\includegraphics[width=0.50\textwidth]{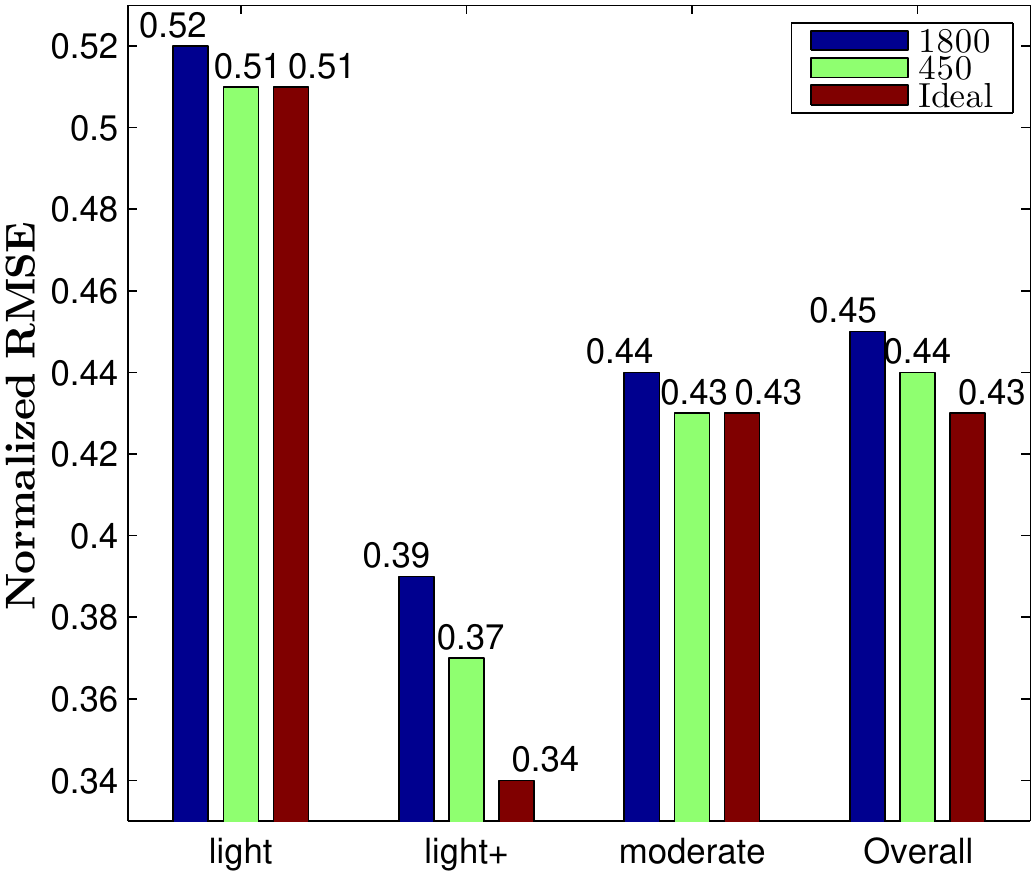}}
\subfloat[]{\includegraphics[width=0.50\textwidth]{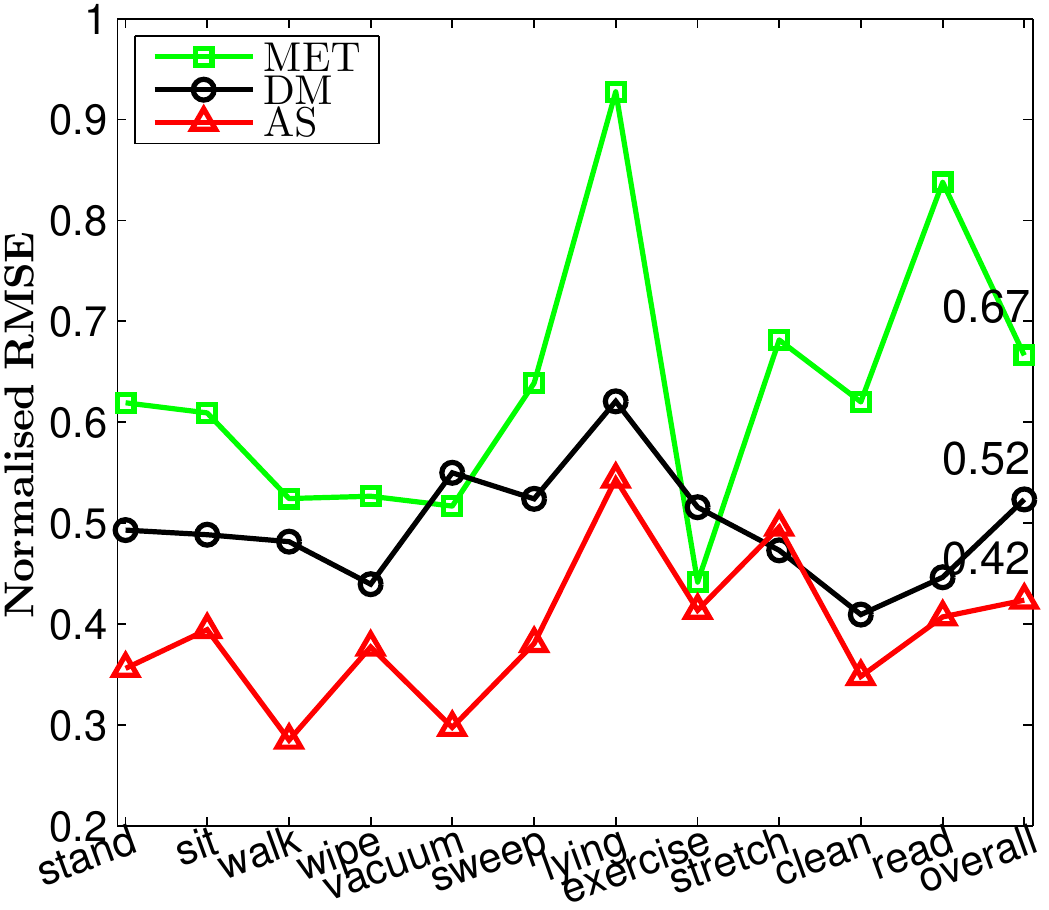}}\vspace{-4pt}
\caption{\textbf{Prediction Accuracy of Calorific Expenditure.} (a) Average calorie prediction errors (Normalised RMSE) for different intensities and action recognition rates; (b) Average calorie prediction errors (Normalised RMSE) of three models.\vspace{-6pt}} 
\label{fig:window_and_models}
\end{figure}

{\bf Evaluation of Recurrent System Layout - }
To evaluate the use of recurrency, we set the activity-specific model using the sliding window {technique above} 
as our baseline method. We now introduce two methods, which are based on recurrent sliding window approaches. The first method {(Recurrent1)} uses the most recent predictions of the baseline method as input together with both visual features to predict current calorie value. Thus, it implements indirect recurrency utilising the predicted values from the baseline as recent predictions. The second method {(Recurrent2)} implements full recurrency, i.e. it uses its own output as recurrent input together with visual features. 

Table \ref{Tab:slidingwindow} shows the effect of using recurrent information, with the best results for each activity highlighted. In general, indirect recurrency, {Recurrent1,} outperforms the other approaches at an average RMSE of 1.24. {We note that the full recurrency, Recurrent2,} suffers from drift and produces the worst results for half of the activities and also overall.

\begin{table*} [t] 
\footnotesize
\begin{center}
\begin{tabular}{|c|c|c|c|c|c|c|c|c|c|c|c|c|}
\hline 
& \textit{stand} & \textit{sit} & \textit{walk} & \textit{wipe} & \textit{vacuum} & \textit{sweep} & \textit{lying} & \textit{exercise} & \textit{stretch} & \textit{clean} & \textit{read} & \textbf{overall} \tabularnewline
\hline 
\hline 
\textbf{Baseline} &  0.70       &0.72   &0.76   &1.09   &1.11   &1.40   &1.51   &2.27   &2.92   &1.43   &1.17   &1.30\tabularnewline
\hline 
\textbf{Recurrent1} &  0.61     &\textbf{0.67}  &\textbf{0.72}  &\textbf{1.01}  &\textbf{1.04}  &\textbf{1.37}  &\textbf{1.44}  &2.11   &2.69   &\textbf{1.37}  &\textbf{1.07}  &\textbf{1.24}\tabularnewline
\hline 
\textbf{Recurrent2} &  \textbf{0.60}    &0.82   &0.80   &1.38   &2.05   &1.91   &1.48   &\textbf{1.95}  &\textbf{2.48}  &1.63   &1.20   &1.50 \tabularnewline
\hline 
\end{tabular}
\caption{Average calorific expenditure prediction errors (RMSE) for each activity with different learning approaches. The best results in each activity are in bold. \vspace{-30pt}} 
\label{Tab:slidingwindow}
\end{center}
\end{table*}

{\bf Model Comparison - }
{We select the indirect recurrency model with the best window configurations as AS and analyse the performance: we compare AS with DM and MET against the ground truth (GT).} For MET, we use the ground truth labels to select activities to keep this procedure identical to the commonly used manual estimate. 
Figure~\ref{fig:window_and_models}(b) shows the average normalised RMSE for each activity estimated by each of the models. Lower values indicate a lower residual. AS gives an average of  0.42 for normalised RMSE, which is 19\% less than DM {at} 0.52 for normalised RMSE, and 36\% less than MET {at} 0.67. The overall improvements are similar across all activities except for upper-body stretching, where DM is slightly better than AS. 

\begin{figure*} [ht]
\centering
\includegraphics[width=1\textwidth]{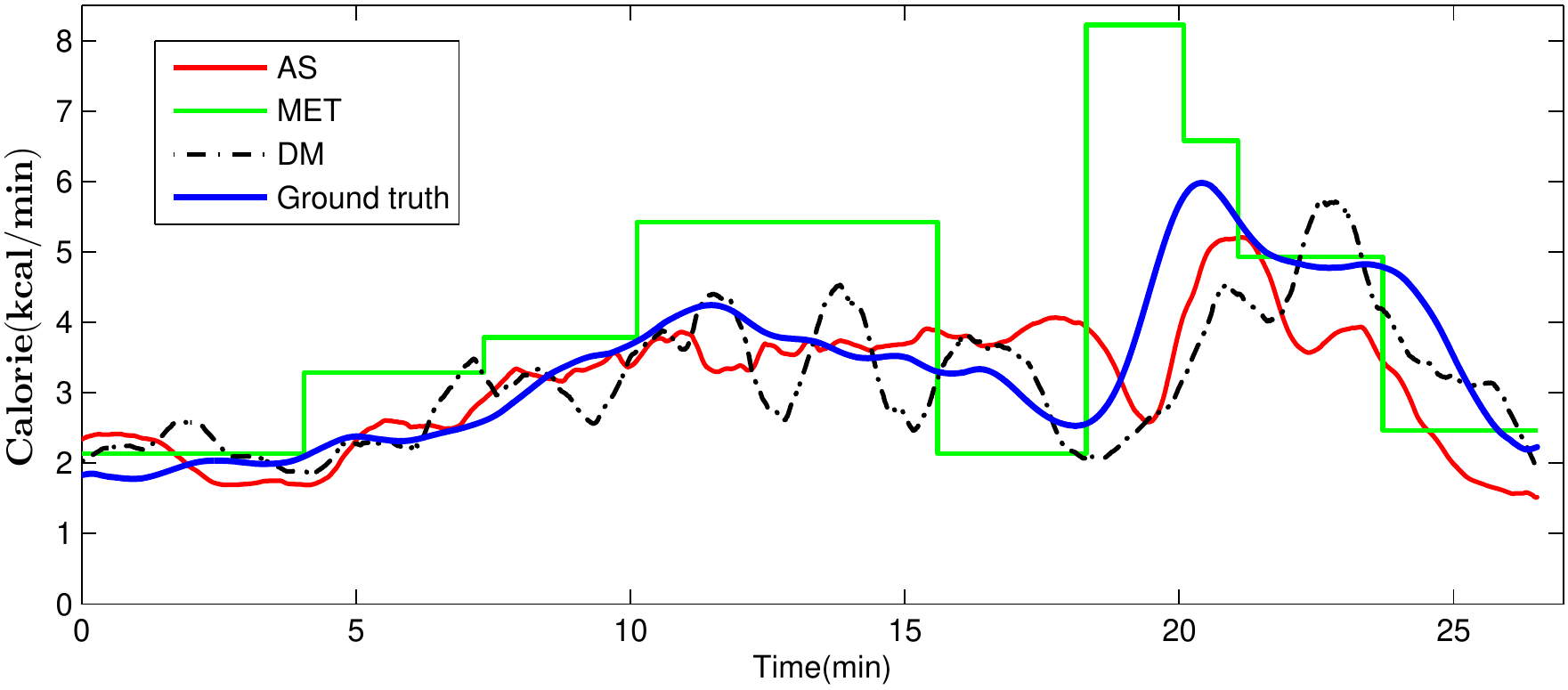} \vspace{-10pt}
\caption{{\bf Example of Calorie Uptake Prediction.} In comparison to DM and  MET, AS shows its ability to better predict  calories and model the transition between activities.\vspace{-10pt}}
\label{Fig:resultmodels}
\end{figure*}

Table \ref{Tab:accuracy} presents the detailed results for each sequence. The accuracy is calculated over the total calorie expended in each recording session. We also measure the correlation between the ground truth and the observed values. Note that the total calorie value for sequence 5, 8, 11, 15 and 16 are relatively low due to {shorter} sequences\footnote{This was caused by camera errors where only the first half of a sequence was saved.}. The proposed AS achieves higher accuracy and correlation in {more} sequences than DM and MET model based methods, {and obtains better rates on average.} 
Figure \ref{Fig:resultmodels} illustrates an example (corresponding to sequence 6 in Table \ref{Tab:accuracy}) of a visual trace of calorie values. 
\\ \ \\ \ \\
The proposed AS matches most closely to the ground truth while DM and MET diverge widely for some time periods. MET, as one of the most commonly used models in many fields,  fails by its nature to capture the transition periods between activities, while the proposed AS model has the ability to capture  transitions fairly well.

\begin{table*} [t]
\footnotesize
\begin{center}
\begin{tabular}{|c|m{1cm}|m{0.7cm}      m{0.7cm}        c|m{1cm}        m{1cm}  m{1cm}  |m{1cm} m{1cm}  m{1cm}|}
\hline 
 & \multicolumn{1}{c|}{} & \multicolumn{3}{c|}{\textbf{Prediction (Calories)}} & \multicolumn{3}{c|}{\textbf{Accuracy \%}} & \multicolumn{3}{c|}{\textbf{Correlation}}\tabularnewline
\hline 
\textbf{sequence} & {GT} & AS & DM & MET & AS & DM & {MET} & AS & DM & MET\tabularnewline
\hline 
\hline 
1 & 59  &71     &83     &76     &\textbf{80.2}  &61.8   &71.3   &\textbf{0.83}  &0.80   &0.66 \tabularnewline
\hline 
2 &  89 &80     &82     &78     &\textbf{90.3}  &90.2   &88.2   &\textbf{0.85}  &0.81   &0.57 \tabularnewline
\hline 
3 &  74 &81     &90     &69     &90.1   &80.7   &\textbf{92.7}  &\textbf{0.84}  &0.78   &0.63 \tabularnewline
\hline 
4 &  79 &48     &46     &43     &\textbf{60.4}  &54.9   &55.0   &\textbf{0.87}  &0.79   &0.78\tabularnewline
\hline 
5 & 37  &39     &41     &28     &\textbf{98.6}  &87.9   &77.6   &\textbf{0.90}  &0.88   &0.77 \tabularnewline
\hline 
6 & 89  &86     &83     &107    &\textbf{94.3}  &93.4   &79.9   &\textbf{0.82}  &0.73   &0.63 \tabularnewline
\hline 
7 & 101 &96     &94     &114    &\textbf{95.3}  &90.1   &87.6 &\textbf{0.61}    &0.57   &0.61\tabularnewline
\hline 
8 &  39 &42     &41     &35     &91.9   &\textbf{94.0}  &84.4   &\textbf{0.93}  &0.42   &0.57\tabularnewline
\hline 
9 &  82 &76     &85     &94     &92.8   &\textbf{96.0}  &85.3   &\textbf{0.87}  &0.78   &0.71 \tabularnewline
\hline 
10 &  49        &68     &77     &76     &\textbf{61.5}  &45.1   &45.2   &0.54   &\textbf{0.61}  &0.42\tabularnewline
\hline 
11 & 28 &38     &40     &38     &65.5   &62.9   &\textbf{66.8}  &\textbf{0.64}  &0.62   &0.56 \tabularnewline
\hline 
12 &  98        &88     &91     &79     &90.3   &\textbf{91.4}  &80.8   &0.56   &0.61   &\textbf{0.66} \tabularnewline
\hline 
13 & 56 &66     &82     &77     &\textbf{82.3}  &55.3&  62.7    &\textbf{0.78}  &0.76   &0.62 \tabularnewline
\hline 
14 & 141        &84     &77     &74     &\textbf{57.4}  &54.9   &52.8   &0.86   &\textbf{0.90}  &0.60 \tabularnewline
\hline 
15 &  40        &41     &40     &30     &\textbf{98.9}  &98.9   &74.5   &\textbf{0.94}  &0.93   &0.94 \tabularnewline
\hline 
16 &  29        &31     &30     &38     &\textbf{97.3}  &94.3   &69.1   &\textbf{0.88}  &0.84   &0.81\tabularnewline
\hline 
17 &  81        &85     &93     &100    &\textbf{94.2}  &88.8   &76.0   &\textbf{0.74}  &0.70   &0.70 \tabularnewline
\hline 
18 &  65        &86     &87     &94     &\textbf{69.3}  &66.1   &54.7   &\textbf{0.83}  &0.70   &0.48 \tabularnewline
\hline 
19 &  92        &89     &84     &101    &\textbf{94.2}  &90.6   &90.6   &0.75   &\textbf{0.81}  &0.72\tabularnewline
\hline 
20 &  63        &83     &82     &86     &66.9   &\textbf{69.5}  &64.4   &\textbf{0.81}  &0.72   &0.41 \tabularnewline
\hline \hline
{\bf Average} & - & - & - & - & \textbf{82.9}&     78.2    &73.7 &  \textbf{0.80}& 0.74&   0.64 \tabularnewline
\hline
\end{tabular}
\caption{Ground truth and predicted calorie values in total per sequence and its accuracy and correlation. The best results for each sequence are in bold. \vspace{-20pt}}
\label{Tab:accuracy}
\end{center}
\end{table*}

\section{Conclusion} 
\vspace{-5pt}This paper presented a system for estimating calorific expenditure from an {RGB-D} sensor. We demonstrated the effectiveness of the method through a comparative study of different models. The proposed activity-specific method used pooled spatial and temporal pyramids of visual features for activity recognition. Subsequently, we utilised a model for each activity built on a recurrent sliding window approach.  To test the methodology, we introduced the challenging \textit{SPHERE-calorie} dataset which covers a wide variety of home-based human activities comprising 20 sequences over 10 subjects. The proposed method demonstrates its ability to outperform the widely used~METs based estimation approach. Possible future work includes taking into account anthropometric features. We hope this {work, and the new dataset,} will establish a baseline for future research in the area. 
\bibliographystyle{splncs}
\bibliography{egbib}


\end{document}